\documentclass{article}
\usepackage{iclr2026_conference,times}

\usepackage{amsmath,amsfonts,bm}

\def\eqref#1{equation~\ref{#1}}

\def\1{\bm{1}}

\DeclareMathAlphabet{\mathsfit}{\encodingdefault}{\sfdefault}{m}{sl}
\SetMathAlphabet{\mathsfit}{bold}{\encodingdefault}{\sfdefault}{bx}{n}

\usepackage{hyperref}
\usepackage{url}
\usepackage{booktabs}
\usepackage{amsmath,amssymb,amsthm}
\usepackage{algorithm}
\usepackage{algorithmic}

\newtheorem{proposition}{Proposition}
\usepackage{graphicx}
\usepackage{subcaption}
\usepackage{xspace}
\usepackage{multirow}

\newcommand{\method}{I-SDPO\xspace}
\newcommand{\sdpo}{SDPO\xspace}
\newcommand{\grpo}{GRPO\xspace}
\newcommand{\srpo}{SRPO\xspace}

\title{\method: Instance-Level Adaptive Self-Distillation\\Policy Optimization}

\author{%
Yubo Zhang \quad Xinhong Ma \quad Zezhong Tan \quad Ziqiang Dong\\
{\normalfont Qwen Large Model Application Team, Alibaba}
}

\iclrfinalcopy

\begin{document}

\maketitle
\pagestyle{plain}

\begin{abstract}
Group Relative Policy Optimization (\grpo) learns from reward differences within a rollout group, but receives no useful relative signal when every sampled response is incorrect. Privileged self-distillation can fill this gap with dense token supervision, yet applying it throughout training creates a different failure mode: the teacher is a biased, low-variance surrogate for the reward objective, so persistent imitation can oppose reward-improving updates after the policy becomes capable of producing successful trajectories. We introduce \method (Instance-Level Adaptive Self-Distillation Policy Optimization), which treats teacher reliance as capability-dependent. \method makes one routing decision per input instance and shares it across that instance's rollout group: all-incorrect groups use a privileged self-distillation objective, whereas any-success groups remain intact for \grpo. This design uses imitation only where group-relative rewards are uninformative. A local analysis characterizes when teacher and reward directions align and shows that a non-vanishing biased distillation weight induces an optimization bias floor. The routing rule automatically reduces the expected distillation rate as success probability rises, withdrawing teacher influence without a hand-designed schedule. On SciKnowEval, \method obtains the best result in all four scientific domains and improves average mean@16 accuracy from 56.67\% with \grpo to 70.31\%, with a maximum domain gain of 18.24 points.
\end{abstract}

\section{Introduction}
\label{sec:intro}

Reinforcement learning (RL) post-training has emerged as a critical paradigm for improving the reasoning capabilities of large language models (LLMs) \citep{ouyang2022training, guo2025deepseekr1}. Among RL approaches, Group Relative Policy Optimization (\grpo) \citep{shao2024deepseekmath} has gained widespread adoption due to its simplicity and effectiveness: it eliminates the need for a separate critic model by computing advantages relative to other samples within the same rollout group.

However, \grpo suffers from a fundamental limitation that we term the \emph{degenerate gradient problem}. When all $K$ responses sampled for a given prompt are incorrect, the within-group advantages collapse to near-zero values, as all rewards are similar. Formally, if rewards $r_1 \approx r_2 \approx \cdots \approx r_K \approx 0$, then the advantage $A_i = r_i - \bar{r} \approx 0$ for all $i$, producing negligible policy gradients. This problem is particularly severe during early training when model capability is low, or on challenging problems where correct solutions are rare---precisely the situations where learning is most needed.

Privileged self-distillation supplies the missing dense signal: a teacher that also observes the ground-truth solution can assign token-level targets along student-generated trajectories, an approach recently instantiated for LLM reasoning by on-policy self-distillation \citep{zhao2026selfdistilled}. This signal has a favorable bias--variance profile early in training. It is biased because matching the teacher distribution is not the same objective as maximizing sequence-level reward, but it has much lower variance than waiting for rare successful samples \citep{menon2021statistical}. Early on, this bias can be preferable to a zero or highly variable RL update.

The same bias becomes consequential later. The privileged teacher is not a reward oracle: it scores tokens under a context containing both the correct solution and the student's possibly erroneous prefix; it may prefer a particular reasoning style even when the reward accepts multiple valid solutions; and, as an exponential moving average (EMA) of the student, it shares the student's systematic errors. Thus, a fixed self-distillation objective can continue pulling toward a teacher-specific distribution after reward-bearing trajectories become available. This is not merely ``noisy supervision.'' It is an objective-mismatch problem whose relative cost grows as the RL estimator becomes more informative. Prior analyses likewise connect distillation to a bias--variance trade-off and show that repeated self-distillation can eventually over-regularize a learner \citep{menon2021statistical,mobahi2020selfdistillation}.

The central question is therefore not whether the teacher is useful in general, but \emph{when it should be trusted}. We propose \method (Instance-Level Adaptive Self-Distillation Policy Optimization), based on a capability-dependent rule: \textbf{make one routing decision per input instance and preserve its complete rollout group}. An all-incorrect group has no relative reward signal and is routed to privileged self-distillation. If at least one response succeeds, the group remains under \grpo; in mixed groups, this preserves the contrast between successful and unsuccessful trajectories. Unlike sample-level routing (\srpo), \method never overwrites the negative samples that make this comparison possible.

This paper makes three contributions:

\begin{itemize}
    \item We formulate privileged self-distillation as a biased, low-variance surrogate for reward optimization. A token-space alignment criterion and a local quadratic analysis explain both its early utility and the bias floor caused by persistent teacher influence.
    \item We introduce instance-level routing over rollout groups, which invokes self-distillation exactly on all-incorrect groups and otherwise preserves group-relative RL. Under conditionally independent sampling, its expected distillation rate is $(1-p)^K$ and therefore decreases automatically with policy success probability $p$.
    \item On four SciKnowEval domains, \method reaches 70.31\% average mean@16 accuracy, outperforming \grpo, pure \sdpo, and sample-level \srpo. Training dynamics and KL-direction ablations are consistent with capability-dependent teacher trust.
\end{itemize}

\section{Related Work}
\label{sec:related}

\paragraph{Reinforcement Learning for LLM Post-Training.}
RL-based post-training traces back to RLHF \citep{ouyang2022training}, which uses Proximal Policy Optimization (PPO) \citep{schulman2017proximal} to align LLMs with human preferences. Direct Preference Optimization (DPO) \citep{rafailov2023direct} simplifies this by eliminating the reward model, directly optimizing preferences via a closed-form objective. More recently, \grpo \citep{shao2024deepseekmath} removes the critic model entirely, computing group-relative advantages from multiple sampled responses. DeepSeek-R1 \citep{guo2025deepseekr1} demonstrates that \grpo-style training can incentivize strong reasoning capabilities. Dr.~GRPO \citep{liu2025drgrpo} identifies and corrects length-dependent biases in the original \grpo formulation. Our work builds on \grpo as the RL backbone and addresses its degenerate gradient problem on all-incorrect rollout groups.

\paragraph{Classical and Online Self-Distillation.}
Knowledge distillation transfers a teacher's predictive distribution or representations to a student \citep{hinton2015distilling}. Self-distillation removes the need for an independently stronger teacher and has taken several forms. Born-Again Networks train successive generations of the same architecture \citep{furlanello2018born}; other methods transfer knowledge between deep and shallow sections of one network \citep{zhang2019byot}, from earlier optimization snapshots \citep{yang2019snapshot}, across distorted views of an example \citep{xu2019datadistortion}, between same-class samples \citep{yun2020classwise}, or from progressively refined past predictions \citep{kim2021progressive}. Online variants construct teachers jointly with the student, using peer networks, native ensembles, or weight-averaged models \citep{zhang2018deep,lan2018one,tarvainen2017mean}. These methods differ in where the target is obtained, but all reuse knowledge already present in the model, training process, or data rather than importing an independent source of task competence.

\paragraph{Theoretical Views of Self-Distillation.}
Several accounts interpret self-distillation primarily as regularization. Teacher predictions can act as instance-specific label smoothing \citep{zhang2020labelsmoothing}; statistically, approximate teacher probabilities trade lower target variance against teacher-induced bias \citep{menon2021statistical}. In a Hilbert-space setting, repeated self-distillation progressively restricts the learned function: a few rounds can reduce overfitting, while further rounds can underfit \citep{mobahi2020selfdistillation}. These results do not directly characterize autoregressive Transformers, but they establish an important qualification to the usual ``dense supervision'' intuition: self-generated targets can help without being unbiased, and continued agreement is not necessarily continued learning.

\paragraph{Distillation for Autoregressive Language Models.}
Sequence-level distillation trains on teacher-generated sequences \citep{kim2016sequence}, whereas recent white-box methods optimize token distributions on student or mixed trajectories. MiniLLM advocates reverse KL for generative distillation \citep{gu2024minillm}; GKD generalizes both the divergence and the use of on-policy student generations \citep{agarwal2024onpolicy}; and DistiLLM combines a skew-KL objective with adaptive off-policy sampling \citep{ko2024distillm}. Token-level teaching behavior is heterogeneous, so applying a single teaching mode uniformly can reduce performance \citep{zhong2024revisiting}. Self-distillation has also been adapted to multilingual transfer and low-bit LLM training \citep{zhang2024multilingual,du2024bitdistiller}, illustrating its breadth beyond conventional model compression.

\paragraph{Privileged On-Policy Self-Distillation and RL.}
Most closely related, Self-Distilled Reasoner introduces on-policy self-distillation (OPSD), where the same LLM serves as a student conditioned on the question and as a teacher additionally conditioned on a verified reasoning trace; the two distributions are matched over student-generated rollouts \citep{zhao2026selfdistilled}. OPSD establishes privileged-context self-distillation as a token-efficient standalone reasoning objective. Our work asks a complementary question exposed by longer training: when should this teacher cease to control optimization? \method couples the privileged objective to verifiable rewards, applying it only to all-incorrect groups and preserving group-relative RL once successful evidence exists. This capability-dependent routing differs from uniform loss blending, pure OPSD, and self-reward or iterative preference schemes \citep{yuan2024selfrewarding,pang2024iterative}.

\section{Method}
\label{sec:method}

\subsection{Preliminaries}
\label{sec:prelim}

\paragraph{Group Relative Policy Optimization (\grpo).}
Given a prompt $x$ and a policy $\pi_\theta$, \grpo samples $K$ responses $\{y_1, \ldots, y_K\}$ and scores each with a reward function $r(x, y_i)$. The advantage for sample $i$ is computed as:
\begin{equation}
    A_i = r(x, y_i) - \frac{1}{K}\sum_{j=1}^{K} r(x, y_j)
\end{equation}
The policy is updated using a clipped surrogate objective following Dr.~GRPO \citep{liu2025drgrpo}:
\begin{equation}
    \mathcal{L}_{\text{GRPO}} = -\sum_{i=1}^{K}\sum_{t=1}^{T_i} \min\!\Big(\rho_{i,t}\, A_i,\; \text{clip}(\rho_{i,t}, 1{-}\epsilon, 1{+}\epsilon)\, A_i\Big)
    \label{eq:grpo}
\end{equation}
where $\rho_{i,t} = \pi_\theta(y_{i,t} \mid x, y_{i,<t}) / \pi_{\theta_{\text{old}}}(y_{i,t} \mid x, y_{i,<t})$ is the per-token importance ratio.

\paragraph{Self-Distillation Policy Optimization (\sdpo).}
\sdpo provides dense token-level supervision by minimizing the divergence between the student policy $\pi_\theta$ and a teacher $\pi_{\text{tea}}$ that observes privileged information. Following on-policy distillation over student-generated trajectories \citep{agarwal2024onpolicy,zhao2026selfdistilled}, we interpolate forward and reverse KL divergences:
\begin{equation}
    \mathcal{L}_{\text{SDPO}} = (1{-}\alpha)\, \text{KL}(\pi_{\text{tea}} \| \pi_\theta)
    + \alpha\, \text{KL}(\pi_\theta \| \pi_{\text{tea}})
    \label{eq:sdpo}
\end{equation}
where $\text{KL}(\pi_{\text{tea}} \| \pi_\theta)$ is the forward KL, $\text{KL}(\pi_\theta \| \pi_{\text{tea}})$ is the reverse KL, and $\alpha \in [0,1]$ controls their balance. The choice matters for autoregressive generation: reverse KL is more mode-seeking, while forward KL places greater weight on teacher coverage \citep{gu2024minillm,ko2024distillm}. Thus, $\alpha{=}0$ uses forward KL only, $\alpha{=}1$ uses reverse KL only, and $\alpha{=}0.5$ weights both directions equally. We use $\alpha{=}0.5$ by default and analyze this choice in Appendix~\ref{app:hyperparams}.

\subsection{The Degenerate Gradient Problem}
\label{sec:degenerate}

We formalize the motivation for \method. Consider a rollout group where all $K$ samples receive identical (or near-identical) rewards $r_i \approx c$ for some constant $c$. The \grpo advantages become:
\begin{equation}
    A_i = r_i - \bar{r} \approx c - c = 0, \quad \forall i \in \{1, \ldots, K\}
\end{equation}
Consequently, $\mathcal{L}_{\text{GRPO}} \approx 0$, and the policy gradient vanishes. With binary rewards, this occurs whenever all responses are incorrect. (An all-correct group is also constant-reward, but it does not represent a failure to discover a successful trajectory.) For a fixed prompt, let $p_t$ denote the per-sample success probability at step $t$. Under conditionally independent sampling, the probability of an all-incorrect group is:
\begin{equation}
    f(t) = (1 - p_t)^K
    \label{eq:all_wrong_prob}
\end{equation}
For $p_t = 0.1$ and $K = 16$, $f(t) \approx 0.185$---nearly one in five groups has no relative reward signal. At $p_t=0.2$ and $0.3$, the same probability falls rapidly to $0.028$ and $0.0033$, respectively. Prompt heterogeneity changes the dataset average to $\mathbb{E}_x[(1-p_t(x))^K]$ but preserves the capability dependence.

\subsection{Why Teacher Trust Should Depend on Capability}
\label{sec:teacher_trust}

Dense supervision is not automatically reward-aligned. At token state $s=(x,y_{<t})$, let $p_s$, $q_s$, and $u_s$ denote the student distribution, privileged-teacher distribution, and a conceptual reward-compatible target distribution. For the forward-KL component, the descent direction with respect to student logits is $q_s-p_s$, whereas an ideal local direction would be $u_s-p_s$. Their alignment is
\begin{equation}
    \Gamma_s=(q_s-p_s)^\top(u_s-p_s)
    =\tfrac{1}{2}\!\left(\lVert q_s-p_s\rVert_2^2+\lVert u_s-p_s\rVert_2^2-\lVert q_s-u_s\rVert_2^2\right).
    \label{eq:teacher_alignment}
\end{equation}
Thus, teacher supervision is locally helpful only when $\Gamma_s>0$. Privileged context can improve $q_s$, but it does not guarantee this condition: the teacher is evaluated after the student's possibly erroneous prefix, and sequence reward may admit many correct continuations that differ from the demonstrated solution. Reverse KL changes the exact logit gradient, but the interpolated objective in Eq.~\ref{eq:sdpo} retains the same dependence on a potentially mismatched $q_s$.

This yields a capability-dependent bias--variance trade-off. Before the policy discovers successful trajectories, the \grpo direction is zero or estimated from rare events; a biased but dense teacher direction can therefore be substantially more useful. Once successful and unsuccessful samples coexist, the group-relative reward supplies an on-policy comparison. The variance-reduction benefit of imitation then shrinks, while any mismatch between $q_s$ and reward-compatible behavior remains. Moreover, teacher and student share an architecture and are coupled by EMA, so agreement may increase either because the student learns or because the teacher inherits the student's errors. EMA smooths temporal fluctuations but cannot remove such shared bias.

The following local model makes the persistent-bias mechanism explicit.
\begin{proposition}[Bias floor under persistent distillation]
\label{prop:bias_floor}
Suppose that near a reward optimum $\theta^\star$, the reward and distillation losses satisfy
$\mathcal{L}_{R}(\theta)\approx\frac{1}{2}\lVert\theta-\theta^\star\rVert_H^2$ and
$\mathcal{L}_{D}(\theta)\approx\frac{1}{2}\lVert\theta-(\theta^\star+b)\rVert_H^2$
for $H\succ0$. Minimizing $\mathcal{L}_{R}+\lambda\mathcal{L}_{D}$ with constant $\lambda\geq0$ gives
\begin{equation}
    \theta_\lambda^\star=\theta^\star+\frac{\lambda}{1+\lambda}b,
    \qquad
    \mathcal{L}_{R}(\theta_\lambda^\star)-\mathcal{L}_{R}(\theta^\star)
    \approx\frac{1}{2}\left(\frac{\lambda}{1+\lambda}\right)^2\lVert b\rVert_H^2.
    \label{eq:bias_floor}
\end{equation}
\end{proposition}

The proposition is a local explanatory model, not a global convergence theorem for Transformers; its proof is in Appendix~\ref{app:theory}. It predicts a bias floor whenever teacher mismatch $b$ and effective distillation weight $\lambda$ both persist. It also explains why pure \sdpo can improve rapidly at first yet plateau or deteriorate later: the supervision that replaces an absent reward gradient early can become the dominant, misaligned force once reward optimization is feasible. Related theory connects distillation to imperfect-teacher bias and repeated self-distillation to progressively stronger regularization \citep{menon2021statistical,mobahi2020selfdistillation}.

\subsection{\method: Instance-Level Routing}
\label{sec:isdpo}

\method addresses the degenerate gradient problem through instance-level routing over rollout groups. Each input instance is a prompt $x_i$, and the routing variable indexed by $i$ is shared by its complete rollout group $\{y_i^1, \ldots, y_i^K\}$. We define:
\begin{align}
    c_i &= \mathbf{1}\!\left[\exists\, j : r(x_i, y_i^j) \geq \tau\right] & &\text{(correctness indicator)} \\
    m_i &= \mathbf{1}\!\left[\text{ground-truth available for } x_i\right] & &\text{(teacher feasibility)}
\end{align}
where $\tau$ is a reward threshold for correctness. The routing masks are:
\begin{align}
    z_i^{\text{SDPO}} &= (1 - c_i) \cdot m_i  \label{eq:z_sdpo} \\
    z_i^{\text{GRPO}} &= 1 - z_i^{\text{SDPO}} \label{eq:z_grpo}
\end{align}

The routing logic is:
\begin{itemize}
    \item If the instance has any correct response ($c_i = 1$): all samples go to \grpo ($z_i^{\text{GRPO}} = 1$); a mixed group therefore retains its reward contrast.
    \item If all responses are wrong and ground-truth exists ($c_i = 0, m_i = 1$): all samples go to \sdpo ($z_i^{\text{SDPO}} = 1$).
    \item If all responses are wrong but no ground-truth ($c_i = 0, m_i = 0$): defaults to \grpo.
\end{itemize}

The combined loss is:
\begin{equation}
    \mathcal{L}_{\text{I-SDPO}} = \frac{\sum_i \left(z_i^{\text{GRPO}} \cdot \mathcal{L}_{\text{GRPO}}^{(i)} + z_i^{\text{SDPO}} \cdot \mathcal{L}_{\text{SDPO}}^{(i)}\right)}{\sum_i \left(z_i^{\text{GRPO}} + z_i^{\text{SDPO}}\right)}
    \label{eq:isdpo_loss}
\end{equation}

\paragraph{Contrast with Sample-Level Routing (\srpo).}
In \srpo, each incorrect sample is routed to \sdpo even when a correct peer exists. Consider a group with 15 incorrect and one correct response. \srpo replaces the learning objective on the 15 negative trajectories, weakening the within-group comparison that identifies which behavior led to success. \method instead keeps the group intact for \grpo and invokes teacher supervision only when the group contains no successful trajectory. The routing decision is therefore also a decision about \emph{teacher trust}: observed policy success makes reward supervision preferable to imitation.

\subsection{Privileged Teacher for Self-Distillation}
\label{sec:teacher}

For instances routed to \sdpo, the teacher model $\pi_{\text{tea}}$ generates token-level soft targets for every rollout in the corresponding group. The teacher shares the student's architecture but receives \emph{privileged information}: the correct solution is prepended to the prompt context before computing log-probabilities over the student's response tokens.

Specifically, for a prompt $x$ with ground-truth solution $y^*$, the teacher input is constructed as:
\begin{equation}
    x_{\text{tea}} = [x \,\|\, \texttt{``Correct solution: ''} \,\|\, y^* \,\|\, \texttt{``Correctly solve the original question.''}]
\end{equation}
The teacher then computes $\pi_{\text{tea}}(\cdot \mid x_{\text{tea}}, y_{<t})$ for the same response tokens $y$ that the student generated. This instantiates privileged on-policy self-distillation \citep{zhao2026selfdistilled} and follows the Learning Using Privileged Information (LUPI) paradigm \citep{vapnik2009learning}: the teacher leverages information available during training but not at inference.

The teacher model is maintained as an Exponential Moving Average (EMA) of the student \citep{tarvainen2017mean}:
\begin{equation}
    \theta_{\text{tea}} \leftarrow (1 - \tau_{\text{ema}})\, \theta_{\text{tea}} + \tau_{\text{ema}}\, \theta_{\text{student}}
    \label{eq:ema}
\end{equation}
where $\tau_{\text{ema}}$ is the update rate. Unrolling Eq.~\ref{eq:ema} shows that past student parameters receive geometrically decaying weights. With $\tau_{\text{ema}}=0.05$, the characteristic averaging horizon is roughly $1/\tau_{\text{ema}}=20$ updates. This smooths short-term target variation, but creates an unavoidable trade-off: faster tracking reduces teacher independence, while slower tracking increases staleness. Neither setting removes systematic teacher--reward mismatch; instance-level routing limits how long that mismatch influences optimization.

\subsection{Entropy-Aware Dynamic Weighting}
\label{sec:entropy}

Not all teacher targets are equally reliable. We use low teacher entropy as a pragmatic confidence proxy and introduce entropy-aware weighting:
\begin{equation}
    w_{i,t} = \frac{\exp(-\beta \cdot H_t^{\text{tea}})}{\frac{1}{|\mathcal{S}|}\sum_{(j,s) \in \mathcal{S}} \exp(-\beta \cdot H_s^{\text{tea}})}
    \label{eq:entropy_weight}
\end{equation}
where $H_t^{\text{tea}} = -\sum_v p_{\text{tea}}(v) \log p_{\text{tea}}(v)$ is the teacher entropy at position $t$, $\beta$ controls sensitivity, and $\mathcal{S}$ contains all \sdpo-routed token positions. Low-entropy positions receive larger weights and high-entropy positions are down-weighted. Importantly, confidence is not correctness: this heuristic reduces exposure to uncertain targets but cannot correct a confidently biased teacher. The weighted \sdpo loss becomes:
\begin{equation}
    \mathcal{L}_{\text{SDPO}}^{(i)} = \sum_{t=1}^{T_i} w_{i,t} \cdot \ell_{\text{KL-mix}}(i, t)
    \label{eq:weighted_sdpo}
\end{equation}
where $\ell_{\text{KL-mix}}(i, t)$ is the per-token forward--reverse KL interpolation from Eq.~\ref{eq:sdpo}.

\subsection{Self-Annealing Property}
\label{sec:annealing}

Instance-level routing converts policy competence into an adaptive effective distillation weight.

\begin{proposition}[Self-Annealing]
\label{prop:annealing}
For a fixed prompt, suppose the $K$ rollouts are conditionally independent with per-sample success probability $p_t$. If $p_t$ is non-decreasing, then the expected \sdpo routing probability $f(t)=(1-p_t)^K$ is non-increasing.
\end{proposition}

The result follows because $(1-p)^K$ decreases in $p$. If the per-routed-group distillation scale is fixed, its expected contribution is nevertheless multiplied by $f(t)$. Hence, as reward-bearing groups become common, \method drives the effective $\lambda$ in Proposition~\ref{prop:bias_floor} toward zero and removes the corresponding bias floor. This capability-coupled withdrawal distinguishes \method from a time-based decay schedule: two prompts at the same training step can receive different supervision according to whether the current policy can solve them.

The complete training procedure is summarized in Algorithm~\ref{alg:isdpo}.

\begin{algorithm}[t]
\caption{\method: Instance-Level Adaptive Self-Distillation Policy Optimization}
\label{alg:isdpo}
\begin{algorithmic}[1]
\REQUIRE Policy $\pi_\theta$, teacher $\pi_{\text{tea}}$, prompts $\mathcal{D}$, reward function $r$, threshold $\tau$, EMA rate $\tau_{\text{ema}}$
\FOR{each training iteration}
    \STATE Sample batch of prompts $\{x_i\}$ from $\mathcal{D}$
    \STATE Generate $K$ responses per prompt: $\{y_i^1, \ldots, y_i^K\} \sim \pi_\theta(\cdot | x_i)$
    \STATE Compute rewards $r(x_i, y_i^j)$ for all samples
    \FOR{each prompt $x_i$}
        \IF{$\exists\, j: r(x_i, y_i^j) \geq \tau$}
            \STATE Route to \grpo: compute $\mathcal{L}_{\text{GRPO}}^{(i)}$ via Eq.~\ref{eq:grpo}
        \ELSIF{ground-truth $y_i^*$ available}
            \STATE Construct teacher input $x_{\text{tea}}$ with $y_i^*$
            \STATE Compute teacher log-probs: $\pi_{\text{tea}}(\cdot | x_{\text{tea}}, y_{i,<t}^j)$
            \STATE Route to \sdpo: compute $\mathcal{L}_{\text{SDPO}}^{(i)}$ via Eq.~\ref{eq:weighted_sdpo}
        \ELSE
            \STATE Route to \grpo (fallback)
        \ENDIF
    \ENDFOR
    \STATE Update $\theta$ using $\mathcal{L}_{\text{I-SDPO}}$ (Eq.~\ref{eq:isdpo_loss})
    \STATE Update teacher: $\theta_{\text{tea}} \leftarrow (1-\tau_{\text{ema}})\theta_{\text{tea}} + \tau_{\text{ema}}\theta$ (Eq.~\ref{eq:ema})
\ENDFOR
\end{algorithmic}
\end{algorithm}

\section{Experiments}
\label{sec:experiments}

\subsection{Experimental Setup}
\label{sec:setup}

\paragraph{Model and Data.}
We use Qwen3-8B \citep{yang2025qwen3} as the base model. Training and evaluation are conducted on the SciKnowEval benchmark \citep{feng2024sciknoweval}, which assesses scientific knowledge across four domains: \textbf{biology}, \textbf{material science}, \textbf{chemistry}, and \textbf{physics}. Each domain provides problems with verifiable answers, enabling rule-based reward computation.

\paragraph{Baselines.}
We compare \method against three baselines:
\begin{itemize}
    \item \textbf{\grpo}: Standard Group Relative Policy Optimization \citep{shao2024deepseekmath} without any self-distillation.
    \item \textbf{\sdpo}: Pure self-distillation applied to all samples without RL, serving as the distillation-only baseline.
    \item \textbf{\srpo}: Sample-level routing where each individual incorrect sample receives \sdpo, even if correct peers exist in the same rollout group.
\end{itemize}

\paragraph{Hyperparameters.}
All methods use a learning rate of $5\times10^{-6}$, forward--reverse KL interpolation $\alpha = 0.5$, teacher EMA update rate $\tau_{\text{ema}} = 0.05$, entropy weighting $\beta = 1.0$, and importance sampling clip ratio of 2.0 for the \sdpo branch. The \grpo branch uses a clip ratio $\epsilon = 0.2$ following Dr.~GRPO \citep{liu2025drgrpo} without standard deviation normalization. Training runs for 2 epochs with 16 rollout samples per prompt. All experiments are conducted using the VERL training framework.
For distillation-based methods, we retain the teacher's 100 highest-probability vocabulary tokens when computing the distillation loss (\texttt{distillation\_topk}$=100$); Appendix~\ref{app:hyperparams} studies the sensitivity to this choice.

\paragraph{Evaluation.}
We report mean@16 accuracy: for each test problem, 16 responses are sampled and the mean accuracy is computed. This metric captures both the model's problem-solving ability and the consistency of its responses.

\subsection{Main Results}
\label{sec:main_results}

\begin{table}[t]
\caption{Main results (mean@16 accuracy, \%) on SciKnowEval across four scientific domains after 2 epochs of training. \textbf{Bold} indicates the best result per domain.}
\label{tab:main}
\begin{center}
\begin{tabular}{lcccc|c}
\toprule
\textbf{Method} & \textbf{Biology} & \textbf{Material} & \textbf{Chemistry} & \textbf{Physics} & \textbf{Avg.} \\
\midrule
\grpo & 32.12 & 70.74 & 62.92 & 60.88 & 56.67 \\
\sdpo & 45.93 & 71.41 & 76.64 & 68.98 & 65.74 \\
\srpo (sample-level) & 44.27 & 72.94 & 78.69 & 68.12 & 66.01 \\
\midrule
\method (instance-level) & \textbf{50.25} & \textbf{74.53} & \textbf{81.16} & \textbf{75.31} & \textbf{70.31} \\
\bottomrule
\end{tabular}
\end{center}
\end{table}

Table~\ref{tab:main} reports final performance. \method is best in every domain and reaches 70.31\% average mean@16 accuracy. It improves over \grpo by 13.64 points on average and by 18.24 points on chemistry, the largest domain-level gain. Pure \sdpo also improves the average over \grpo, from 56.67\% to 65.74\%, showing that dense privileged supervision is valuable when reward-only learning is sparse. Its 4.57-point gap behind \method is equally informative: teacher supervision is useful, but applying it indiscriminately is not the best final objective.

Routing granularity accounts for a further difference. \method exceeds sample-level \srpo in biology, material science, chemistry, and physics by 5.98, 1.59, 2.47, and 7.19 points, respectively, for a 4.30-point average gain. These results are consistent with preserving mixed groups for reward learning: once a group contains both outcomes, its incorrect trajectories are valuable negative evidence rather than merely samples to be imitated under privileged context.

\subsection{Training Dynamics Analysis}
\label{sec:dynamics}

We next examine whether the optimization dynamics match the capability-dependent account in Section~\ref{sec:teacher_trust}.

\begin{figure}[t]
\begin{center}
\begin{subfigure}[b]{0.48\linewidth}
    \includegraphics[width=\linewidth]{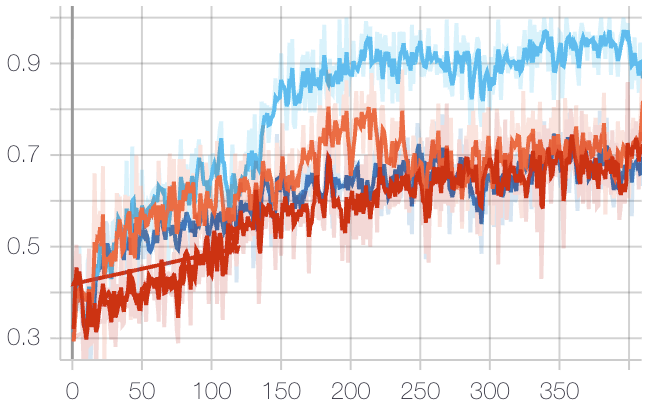}
    \caption{Biology}
\end{subfigure}
\hfill
\begin{subfigure}[b]{0.48\linewidth}
    \includegraphics[width=\linewidth]{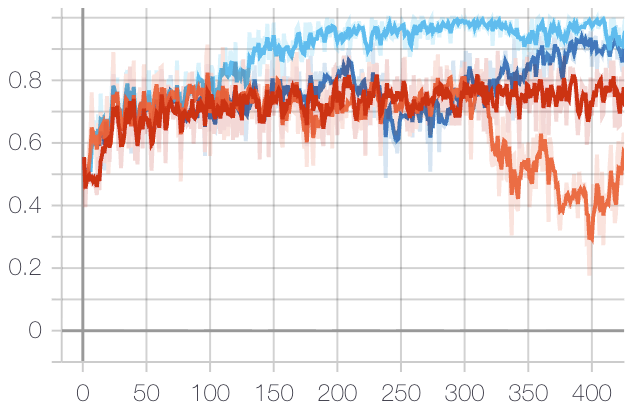}
    \caption{Material Science}
\end{subfigure}
\\[0.5em]
\begin{subfigure}[b]{0.48\linewidth}
    \includegraphics[width=\linewidth]{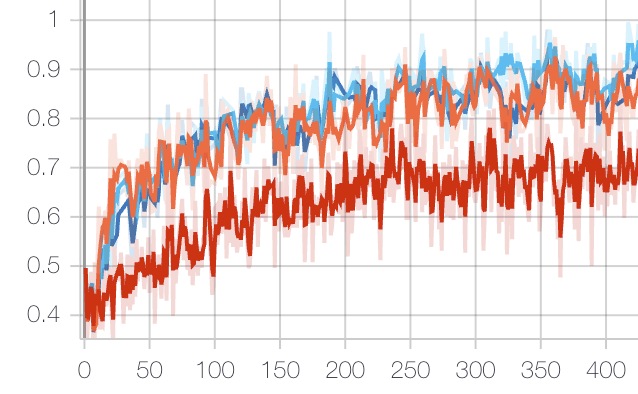}
    \caption{Chemistry}
\end{subfigure}
\hfill
\begin{subfigure}[b]{0.48\linewidth}
    \includegraphics[width=\linewidth]{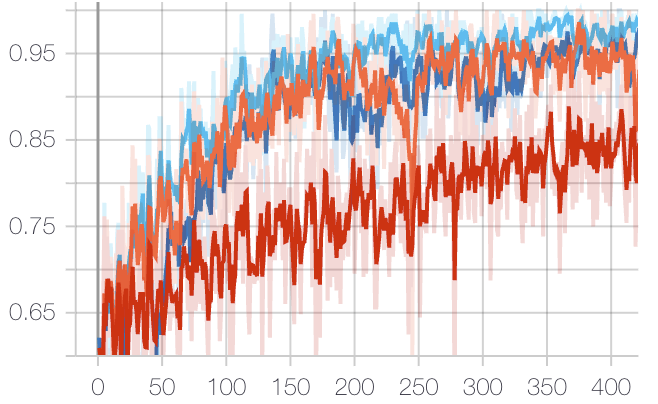}
    \caption{Physics}
\end{subfigure}
\end{center}
\caption{Mean training reward across SciKnowEval domains. Distillation-based methods improve rapidly in the early phase, when reward-bearing rollouts are scarce; \method retains this bootstrapping effect while reducing teacher use as the policy improves.}
\label{fig:reward_dynamics}
\end{figure}

\paragraph{Reward Curves.}
Figure~\ref{fig:reward_dynamics} shows mean training reward. Across domains, methods containing self-distillation rise faster than pure \grpo early in training. This is the regime in which all-incorrect groups are common and token-level targets replace otherwise absent relative-reward updates. Final accuracy in Table~\ref{tab:main}, however, favors \method over pure \sdpo. Together, the observations support a phase-dependent interpretation: dense imitation is most useful for bootstrapping, whereas reward optimization should dominate after successful trajectories become available.

\begin{figure}[t]
\begin{center}
\begin{subfigure}[b]{0.24\linewidth}
    \includegraphics[width=\linewidth]{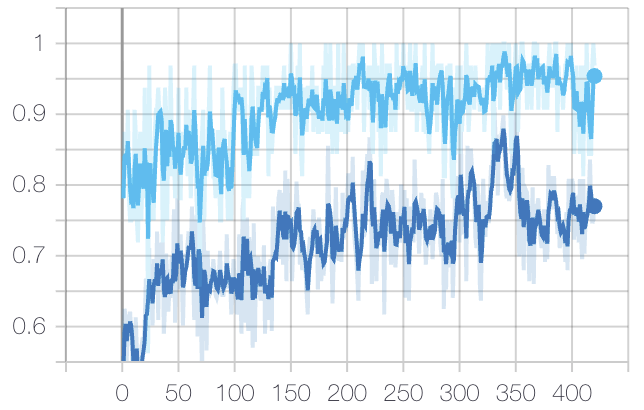}
    \caption{Biology}
\end{subfigure}
\hfill
\begin{subfigure}[b]{0.24\linewidth}
    \includegraphics[width=\linewidth]{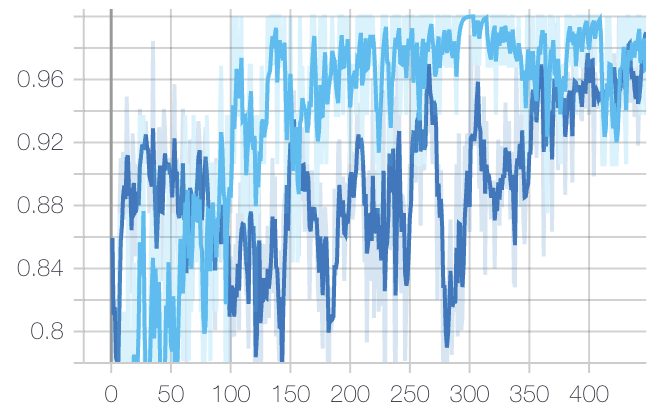}
    \caption{Material}
\end{subfigure}
\hfill
\begin{subfigure}[b]{0.24\linewidth}
    \includegraphics[width=\linewidth]{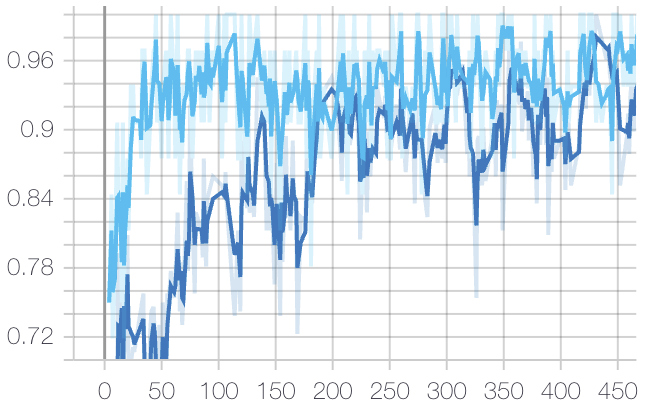}
    \caption{Chemistry}
\end{subfigure}
\hfill
\begin{subfigure}[b]{0.24\linewidth}
    \includegraphics[width=\linewidth]{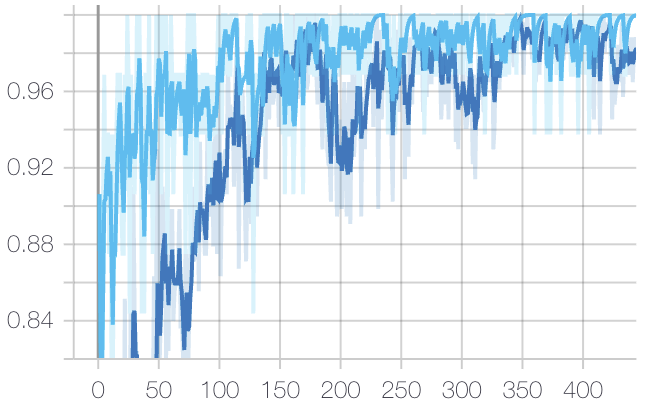}
    \caption{Physics}
\end{subfigure}
\\[0.5em]
\begin{subfigure}[b]{0.24\linewidth}
    \includegraphics[width=\linewidth]{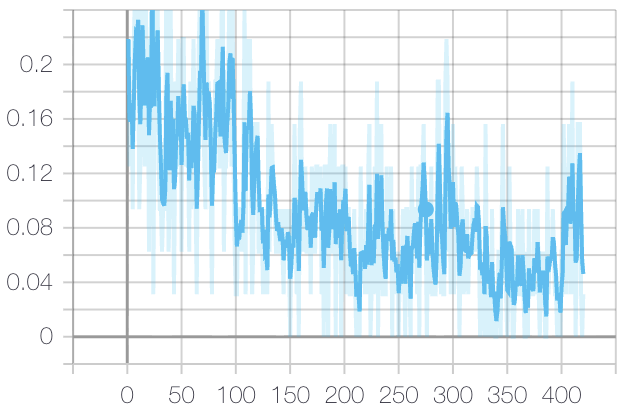}
    \caption{Biology}
\end{subfigure}
\hfill
\begin{subfigure}[b]{0.24\linewidth}
    \includegraphics[width=\linewidth]{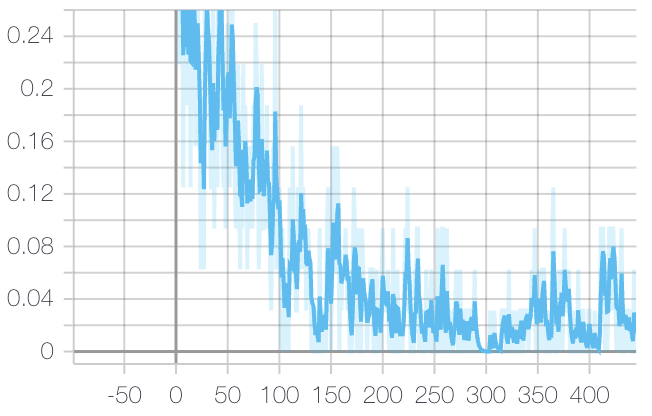}
    \caption{Material}
\end{subfigure}
\hfill
\begin{subfigure}[b]{0.24\linewidth}
    \includegraphics[width=\linewidth]{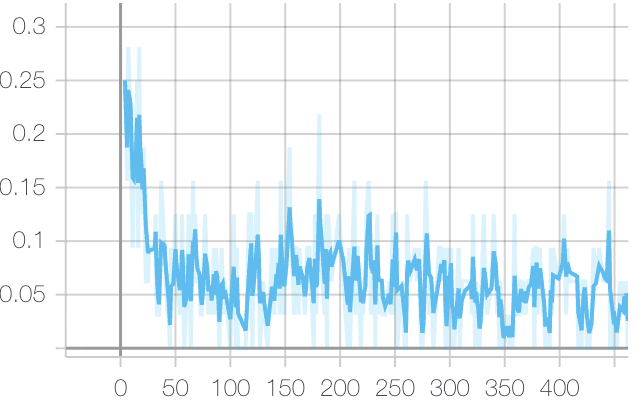}
    \caption{Chemistry}
\end{subfigure}
\hfill
\begin{subfigure}[b]{0.24\linewidth}
    \includegraphics[width=\linewidth]{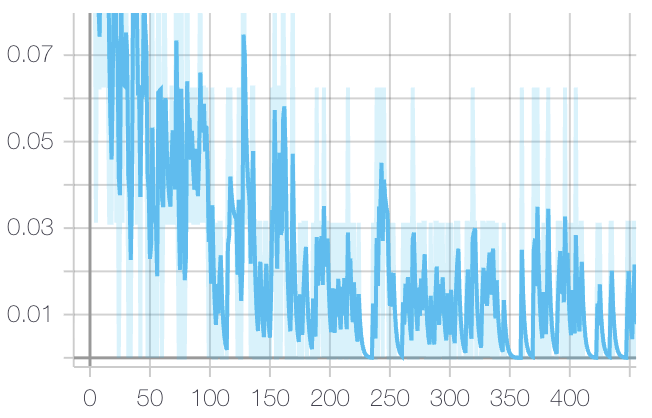}
    \caption{Physics}
\end{subfigure}
\end{center}
\caption{Capability-dependent routing in \method. (Top) The fraction assigned to \grpo generally increases during training. (Bottom) The all-wrong group fraction correspondingly decreases, so teacher influence is withdrawn as successful sampling becomes more likely.}
\label{fig:annealing}
\end{figure}

\paragraph{Self-Annealing Behavior.}
Figure~\ref{fig:annealing} shows the fraction of samples routed to \grpo and the all-wrong group fraction. The \grpo fraction trends upward (from approximately 0.78 to 0.96 on physics and 0.65 to 0.90 on biology), while all-wrong groups become less frequent. This is the feedback predicted by Proposition~\ref{prop:annealing}: improved sampling success directly reduces the effective weight of the biased surrogate rather than relying on elapsed training time.

\paragraph{Lower Intrusion on Moderate-Difficulty Data.}
A notable difference from \srpo is that \method assigns a larger fraction to \grpo on material science and physics (approximately 0.90--0.96 versus 0.82--0.90). The stronger final results despite less distillation coverage argue against ``more teacher supervision is always better.'' They instead favor selective intervention: keep the low-variance teacher for groups with no reward contrast, and preserve on-policy evidence everywhere else.

\begin{figure}[t]
\begin{center}
\begin{subfigure}[b]{0.48\linewidth}
    \includegraphics[width=\linewidth]{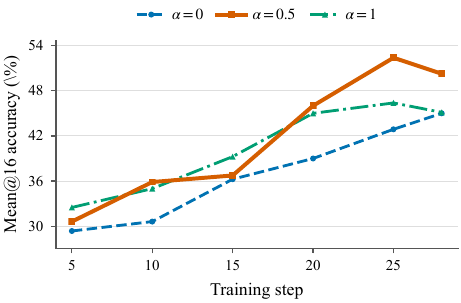}
    \caption{Biology}
\end{subfigure}
\hfill
\begin{subfigure}[b]{0.48\linewidth}
    \includegraphics[width=\linewidth]{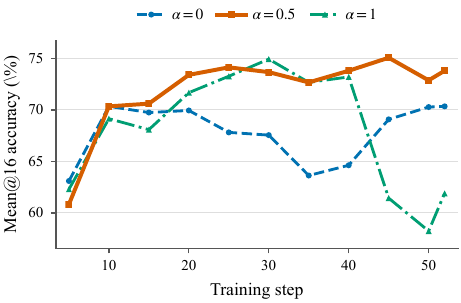}
    \caption{Material Science}
\end{subfigure}
\\[0.4em]
\begin{subfigure}[b]{0.48\linewidth}
    \includegraphics[width=\linewidth]{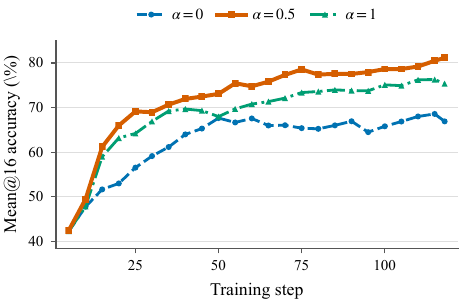}
    \caption{Chemistry}
\end{subfigure}
\hfill
\begin{subfigure}[b]{0.48\linewidth}
    \includegraphics[width=\linewidth]{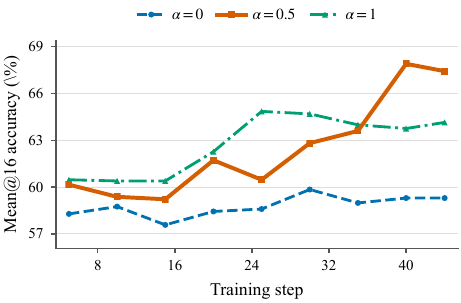}
    \caption{Physics}
\end{subfigure}
\end{center}
\caption{Effect of the forward--reverse KL interpolation on \method across four SciKnowEval domains. $\alpha=0$ uses forward KL only, $\alpha=1$ uses reverse KL only, and $\alpha=0.5$ weights both directions equally. Curves show the raw mean@16 evaluation accuracy recorded during training.}
\label{fig:alpha_ablation}
\end{figure}

\paragraph{Forward--Reverse KL Ablation.}
Figure~\ref{fig:alpha_ablation} compares mean@16 accuracy trajectories for $\alpha\in\{0,0.5,1\}$ in Eq.~\ref{eq:sdpo}. The balanced objective ($\alpha=0.5$) has the highest final accuracy in all four domains. This is consistent with combining the mode-covering tendency of forward KL and the sharper mode preference of reverse KL, although balancing divergence directions does not by itself remove teacher--reward mismatch. Appendix~\ref{app:hyperparams} gives the quantitative results.

\subsection{Capability-Dependent Teacher Trust}
\label{sec:sweet_spot}

The theory and results suggest a single organizing principle: the correct amount of teacher trust is determined by the policy's current ability to produce endogenous reward supervision. At low capability, \grpo often observes no outcome variation; the teacher then supplies a biased but low-variance direction. At higher capability, mixed groups expose which sampled behaviors succeed, while the privileged teacher remains tied to a particular solution and to the student's own EMA history. The signal-to-bias ratio of distillation therefore falls even if the teacher becomes numerically closer to the student.

This view explains why a fixed \sdpo objective can help most at the beginning yet become limiting later. Student--teacher disagreement is not a stationary measure of teaching value: easy, aligned disagreements are absorbed first, leaving residual KL pressure concentrated on difficult prefixes, alternative valid modes, and confidently miscalibrated targets. Continued minimization can then reduce KL without improving sequence reward. This mechanism complements views of self-distillation as instance-specific smoothing and progressively stronger regularization \citep{zhang2020labelsmoothing,mobahi2020selfdistillation}. EMA delays and smooths this process, but cannot create independent knowledge or correct shared error.

\method operationalizes the principle without estimating gradient alignment directly. The binary event ``no successful response among $K$ samples'' is an observable proxy for whether relative reward is informative. Its probability $(1-p)^K$ falls steeply with capability, so the algorithm moves from teacher-led bootstrapping to reward-led refinement at an instance-dependent rate. Instance-level routing matters because the decision is shared by all rollouts of the same input: whenever successful evidence exists, the complete group comparison is preserved, whereas sample-level routing partially discards it.

\section{Discussion}
\label{sec:discussion}

\paragraph{Connection to Curriculum Learning.}
\method resembles a curriculum \citep{bengio2009curriculum}, but the curriculum is over \emph{supervision sources}, not a fixed ordering of examples. The same prompt receives dense privileged supervision when the current policy fails to sample any success and group-relative feedback once it does. Difficulty is therefore model- and time-dependent.

\paragraph{Connection to Learning Using Privileged Information.}
The teacher follows Learning Using Privileged Information (LUPI) \citep{vapnik2009learning}: ground-truth solutions are available during training but not inference. Our analysis adds an important qualification. Privilege can make targets more informative without making them identical to the reward-optimal policy, especially when the teacher evaluates an erroneous student prefix. \method therefore uses privilege as a temporary scaffold rather than a permanent optimization target.

\paragraph{Scope of the theory.}
Equation~\ref{eq:teacher_alignment} is an exact local statement for the forward-KL logit direction, while Proposition~\ref{prop:bias_floor} is a quadratic approximation around a reward optimum. They identify concrete mechanisms---directional mismatch, persistent bias, and capability-dependent variance reduction---but do not establish global convergence or predict every non-convex training trajectory. The empirical dynamics should consequently be read as consistent with the account, not as proof of its assumptions.

\section{Limitations}
\label{sec:limitations}

We acknowledge two limitations:

\textbf{Model scale.} Our experiments use only Qwen3-8B. Results on other model families and scales may differ, particularly because stronger base policies change the frequency and duration of all-incorrect rollout groups.

\textbf{Benchmark diversity.} Evaluation is limited to SciKnowEval. The relative value of privileged distillation and group-relative rewards may change on general mathematical or open-ended reasoning tasks with different reward structure and solution multiplicity.

\section{Conclusion}
\label{sec:conclusion}

We presented \method, which treats privileged self-distillation as a capability-dependent scaffold rather than a uniformly reliable target. The central distinction is a bias--variance one: distillation supplies a dense, low-variance direction when all sampled rewards are uninformative, but its teacher-specific bias can obstruct reward optimization after the policy begins producing successful trajectories. Instance-level routing uses the existence of such trajectories to choose one objective for the corresponding rollout group and automatically withdraws the teacher as capability improves.

On SciKnowEval, \method achieves the best result in all four domains and 70.31\% average mean@16 accuracy, improving over \grpo, pure \sdpo, and sample-level \srpo by 13.64, 4.57, and 4.30 points, respectively. The reward and routing dynamics are consistent with the proposed transition from teacher-led bootstrapping to reward-led refinement. More broadly, our results indicate that the value of self-distillation lies not only in the quality of its targets, but in restricting those targets to the phase in which they are preferable to the policy's own reward evidence.

\subsubsection*{Reproducibility Statement}
We provide full hyperparameter specifications in Section~\ref{sec:setup}. The method description in Section~\ref{sec:method} and Algorithm~\ref{alg:isdpo} contain sufficient detail for re-implementation. Code will be released upon publication.

\bibliography{iclr2026_conference}
\bibliographystyle{iclr2026_conference}

\appendix
\section{Analysis Details}
\label{app:theory}

\paragraph{Token-direction identity.}
Let $z_s$ be the student logits and $p_s=\operatorname{softmax}(z_s)$. For the forward-KL term, $\nabla_{z_s}\operatorname{KL}(q_s\|p_s)=p_s-q_s$; hence its negative-gradient direction is $q_s-p_s$. Replacing $q_s$ with a conceptual reward-compatible target $u_s$ gives $u_s-p_s$. Equation~\ref{eq:teacher_alignment} then follows from the polarization identity
$2a^\top b=\lVert a\rVert_2^2+\lVert b\rVert_2^2-\lVert a-b\rVert_2^2$
with $a=q_s-p_s$ and $b=u_s-p_s$.

\paragraph{Proof of Proposition~\ref{prop:bias_floor}.}
Under the stated local approximations, the gradient of the combined objective is
\begin{equation}
    \nabla_\theta(\mathcal{L}_R+\lambda\mathcal{L}_D)
    \approx H(\theta-\theta^\star)+\lambda H(\theta-\theta^\star-b).
\end{equation}
Setting this expression to zero and using $H\succ0$ yields
$(1+\lambda)(\theta-\theta^\star)=\lambda b$, and therefore
$\theta_\lambda^\star=\theta^\star+\frac{\lambda}{1+\lambda}b$.
Substitution into the quadratic approximation of $\mathcal{L}_R$ gives Eq.~\ref{eq:bias_floor}. In particular, the excess reward loss is positive whenever $b\neq0$ and $\lambda>0$, and vanishes as routing drives the effective $\lambda$ to zero.

\paragraph{EMA memory.}
Unrolling Eq.~\ref{eq:ema} at update $t$ gives
\begin{equation}
    \theta_{\mathrm{tea},t}=(1-\tau_{\mathrm{ema}})^t\theta_{\mathrm{tea},0}
    +\tau_{\mathrm{ema}}\sum_{j=1}^{t}(1-\tau_{\mathrm{ema}})^{t-j}\theta_{\mathrm{student},j}.
\end{equation}
Ignoring the vanishing initialization term, these weights form a geometric age distribution with mean age $(1-\tau_{\mathrm{ema}})/\tau_{\mathrm{ema}}$. For $\tau_{\mathrm{ema}}=0.05$, the mean age is 19 updates and the half-life is $\log(1/2)/\log(0.95)\approx13.5$ updates. This quantifies smoothing and lag; it does not imply that EMA corrects systematic errors.

\section{Detailed Experimental Results}
\label{app:detailed}

Table~\ref{tab:detailed} provides a detailed view of the training dynamics metrics at different training checkpoints.

\begin{table}[h]
\caption{Training dynamics at selected checkpoints for \method on biology.}
\label{tab:detailed}
\begin{center}
\begin{tabular}{lccc}
\toprule
\textbf{Step} & \textbf{GRPO Fraction} $\uparrow$ & \textbf{All-Wrong Fraction} $\downarrow$ & \textbf{Mean Reward} $\uparrow$ \\
\midrule
50 & $\sim$0.65 & $\sim$0.22 & $\sim$0.20 \\
150 & $\sim$0.78 & $\sim$0.12 & $\sim$0.50 \\
250 & $\sim$0.85 & $\sim$0.06 & $\sim$0.65 \\
350 & $\sim$0.90 & $\sim$0.03 & $\sim$0.72 \\
\bottomrule
\end{tabular}
\end{center}
\end{table}

\section{Hyperparameter Sensitivity}
\label{app:hyperparams}

\paragraph{Distillation top-$k$.}
This experiment tests whether the performance of instance-level and sample-level routing depends on the number of teacher vocabulary tokens retained by the distillation objective. We vary \texttt{distillation\_topk} over $\{20,40,60,80,100\}$ while keeping all other training settings fixed. Table~\ref{tab:distillation_topk} reports the resulting mean@16 accuracy.

\begin{table}[h]
\caption{Sensitivity to \texttt{distillation\_topk} for \method and \srpo on SciKnowEval (mean@16 accuracy, \%). \textbf{Bold} indicates the best top-$k$ value within each row. The average is computed across the four domains.}
\label{tab:distillation_topk}
\begin{center}
\begin{tabular}{llccccc}
\toprule
\textbf{Method} & \textbf{Domain} & \textbf{20} & \textbf{40} & \textbf{60} & \textbf{80} & \textbf{100} \\
\midrule
\multirow{5}{*}{\method}
& Biology   & 50.13 & \textbf{54.12} & 53.25 & 47.00 & 50.25 \\
& Material  & 75.44 & \textbf{76.51} & 75.37 & 74.83 & 74.53 \\
& Chemistry & 79.49 & 78.13 & 77.92 & 77.21 & \textbf{81.16} \\
& Physics   & 71.09 & 62.42 & 72.89 & 67.31 & \textbf{75.31} \\
\cmidrule(lr){2-7}
& \textit{Average} & 69.04 & 67.80 & 69.86 & 66.59 & \textbf{70.31} \\
\midrule
\multirow{5}{*}{\srpo}
& Biology   & 46.88 & 46.13 & 54.25 & \textbf{55.87} & 44.27 \\
& Material  & 71.24 & \textbf{74.64} & 72.91 & 72.02 & 72.94 \\
& Chemistry & 77.34 & 77.35 & 78.67 & 76.92 & \textbf{78.69} \\
& Physics   & 53.67 & 67.27 & 60.16 & 55.78 & \textbf{68.12} \\
\cmidrule(lr){2-7}
& \textit{Average} & 62.28 & 66.35 & \textbf{66.50} & 65.15 & 66.01 \\
\bottomrule
\end{tabular}
\end{center}
\end{table}

Neither routing method improves monotonically with top-$k$. For \method, $k=40$ performs best on biology and material science, whereas $k=100$ performs best on chemistry, physics, and the cross-domain average; \srpo shows a similarly mixed pattern and peaks on average at $k=60$. Smaller values remain competitive, so retaining more teacher tokens does not necessarily strengthen distillation. We hypothesize that many instances are easy enough for the teacher to concentrate task-relevant mass on a small token set; larger $k$ then mainly adds a low-probability tail. Because we did not measure retained probability mass directly, this remains a hypothesis. We use $k=100$ in the main experiments.

\paragraph{Forward--reverse KL balance.}
We compare $\alpha\in\{0,0.5,1\}$ in Eq.~\ref{eq:sdpo} with the remaining settings fixed to test whether distillation benefits from balancing the mode-covering behavior of forward KL and the mode-seeking behavior of reverse KL. Figure~\ref{fig:alpha_ablation} reports the raw TensorBoard mean@16 measurements.

The balanced objective ($\alpha=0.5$) achieves the highest final accuracy in every domain: 50.25\% on biology, 73.80\% on material science, 81.16\% on chemistry, and 67.42\% on physics. Its 68.16\% cross-domain average exceeds forward KL alone (60.39\%) and reverse KL alone (61.63\%). The two directions are complementary: forward KL promotes coverage, whereas reverse KL sharpens high-probability modes but can become overly mode-seeking. Because each curve is a single run, we compare their recorded final values and treat trajectory details descriptively.

\end{document}